# Adaptive data selection improves wearable prediction under low baseline performance

Ali Kargarandehkordi, PhD

## Abstract

Adaptive sensing strategies that selectively sample data are increasingly used in wearable health systems to improve prediction performance under limited data budgets, yet their benefits across individuals remain poorly understood. Here, we evaluate adaptive selection of time windows for model training under fixed measurement budgets across multiple sensing modalities, including heart rate, activity, and ecological momentary assessment (EMA), in a longitudinal wearable dataset. We quantify performance gains relative to random sampling using both area under the receiver operating characteristic curve (AUROC) and F1 score. Adaptive strategies yield substantial improvements in AUROC for participants with low baseline performance (with gains up to 0.7), while offering limited or negative gains for participants with strong baselines. Across modalities, adaptive gain is strongly inversely correlated with baseline performance (Pearson $r = -0.67$; Spearman $\rho = -0.62$; $p < 10^{-30}$). At the participant level, most individuals benefit in AUROC (60–80% across modalities), although improvements in F1 are smaller and less consistent. These findings show that adaptive sensing is not uniformly beneficial, but instead provides the greatest value in underperforming settings. Our results support selective deployment strategies that tailor adaptive sensing based on baseline performance to improve efficiency in wearable health monitoring.

## Introduction

Wearable devices enable continuous monitoring of physiological and behavioral signals, creating new opportunities for real-time health prediction and intervention [1-3]. However, collecting high-frequency multimodal data in real-world settings remains challenging due to device limitations, user burden, and adherence constraints [4-6]. This has led to growing interest in adaptive sensing strategies that selectively acquire data at informative time points to improve predictive performance while reducing measurement cost [7-9].

Adaptive sensing and active learning approaches have shown promise across a range of digital health applications, including stress detection, mental health monitoring, and cardiovascular risk prediction [10-12]. These methods typically prioritize data acquisition based on model uncertainty, predicted risk, or hybrid criteria, aiming to maximize information gain under limited sensing budgets [7, 8]. Despite their increasing adoption, it remains unclear when adaptive sensing provides meaningful benefits in wearable health systems, particularly in heterogeneous, real-world populations [13].

A key challenge is that wearable data are often highly redundant and multimodal, combining physiological signals, behavioral context, and self-reported measures [14]. Prior work has demonstrated that integrating passive sensing with EMA can improve the detection of substance use, stress, and mental health states, but also introduces variability due to adherence and reporting behavior [4,5,15-17]. In such settings, random or uniform sampling may already capture sufficient information, potentially limiting the advantage of adaptive strategies. Conversely, in low-signal or sparse sensing conditions, adaptive policies may play a more critical role by prioritizing informative observations [10, 11]. However, the extent to which adaptive sensing benefits depend on signal richness and individual-level characteristics has not been systematically evaluated.

Recent advances in personalized and self-supervised modeling show that accounting for individual variability is critical in wearable health systems [12, 18]. Personalized models can capture subject-specific physiological patterns, improving prediction accuracy for outcomes such as stress and blood pressure fluctuations [12, 19]. These findings suggest that the effectiveness of adaptive sensing may also vary across individuals, depending on baseline model performance and data quality.

n this study, we investigate whether model-guided selection of training observations can improve wearable health prediction under constrained data budgets. Rather than training models on all available temporal windows, we simulate scenarios in which only a subset of candidate observation windows is retained for model training. Each observation window contains wearable and contextual features derived from heart rate, physical activity, ecological momentary assessment (EMA), and temporal context. We compare random window selection against adaptive data selection strategies that prioritize windows based on predictive uncertainty, predicted risk, or hybrid scoring approaches. Under this formulation, adaptive data selection refers to selecting informative training windows rather than uniformly sampling windows at random. Importantly, the sensing modalities themselves remain unchanged across strategies; only the subset of training observations differs. We evaluate these policies across multiple sensing configurations and training data budgets using AUROC and F1 score.

## Results

### Adaptive sensing performance varies with baseline model strength

We first examined the relationship between baseline performance under random window selection and the improvement achieved through adaptive data selection. For each participant and budget, we report the best-performing adaptive strategy to summarize the upper bound of adaptive policy performance. Across all modality configurations, we observed a consistent negative association between baseline performance and adaptive gain (Figures 1, 2). For each participant and budget, adaptive gain was defined relative to the corresponding random window selection baseline.

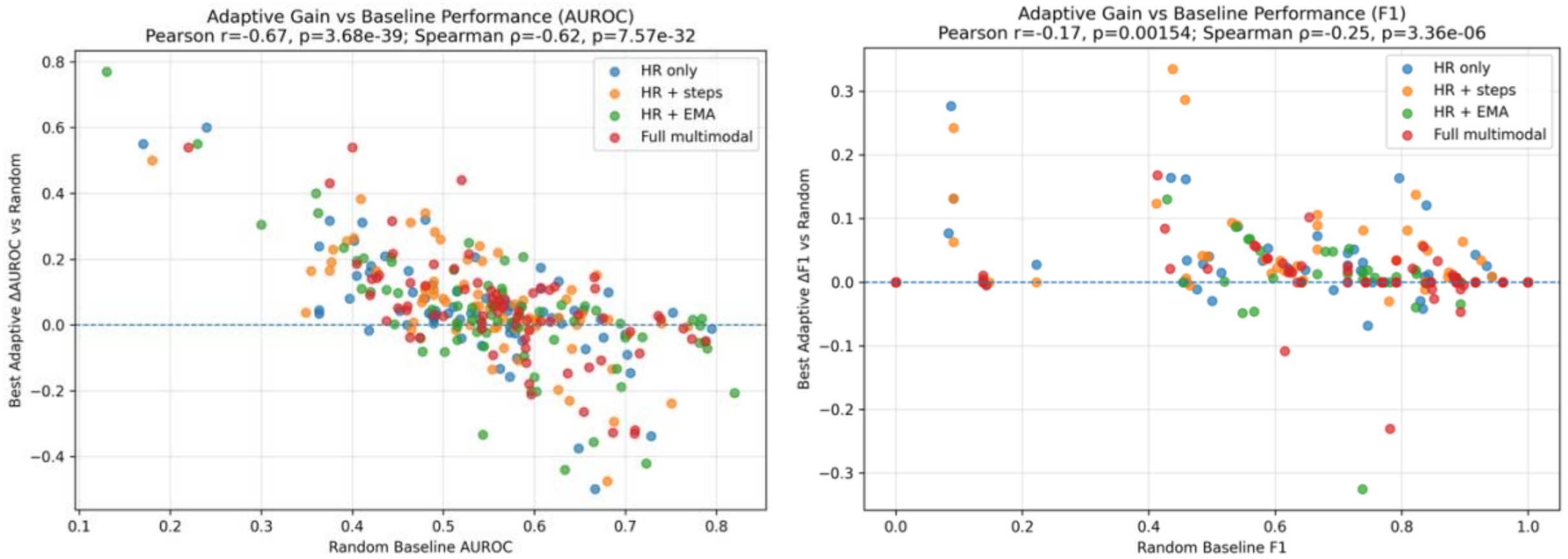

**Figure 1. Adaptive gain as a function of baseline performance.**
(Left) Relationship between baseline model performance (random sampling) and adaptive gain (AUROC). Each point represents a participant–modality–budget combination. A strong negative correlation is observed (Pearson $r = -0.67$; Spearman $\rho = -0.62$; $p < 1e-30$), indicating that adaptive sensing provides the highest benefit for participants with low baseline performance. (Right) Same analysis for F1 score, showing a weaker but significant negative relationship (Pearson $r = -0.17$; Spearman $\rho = -0.25$).

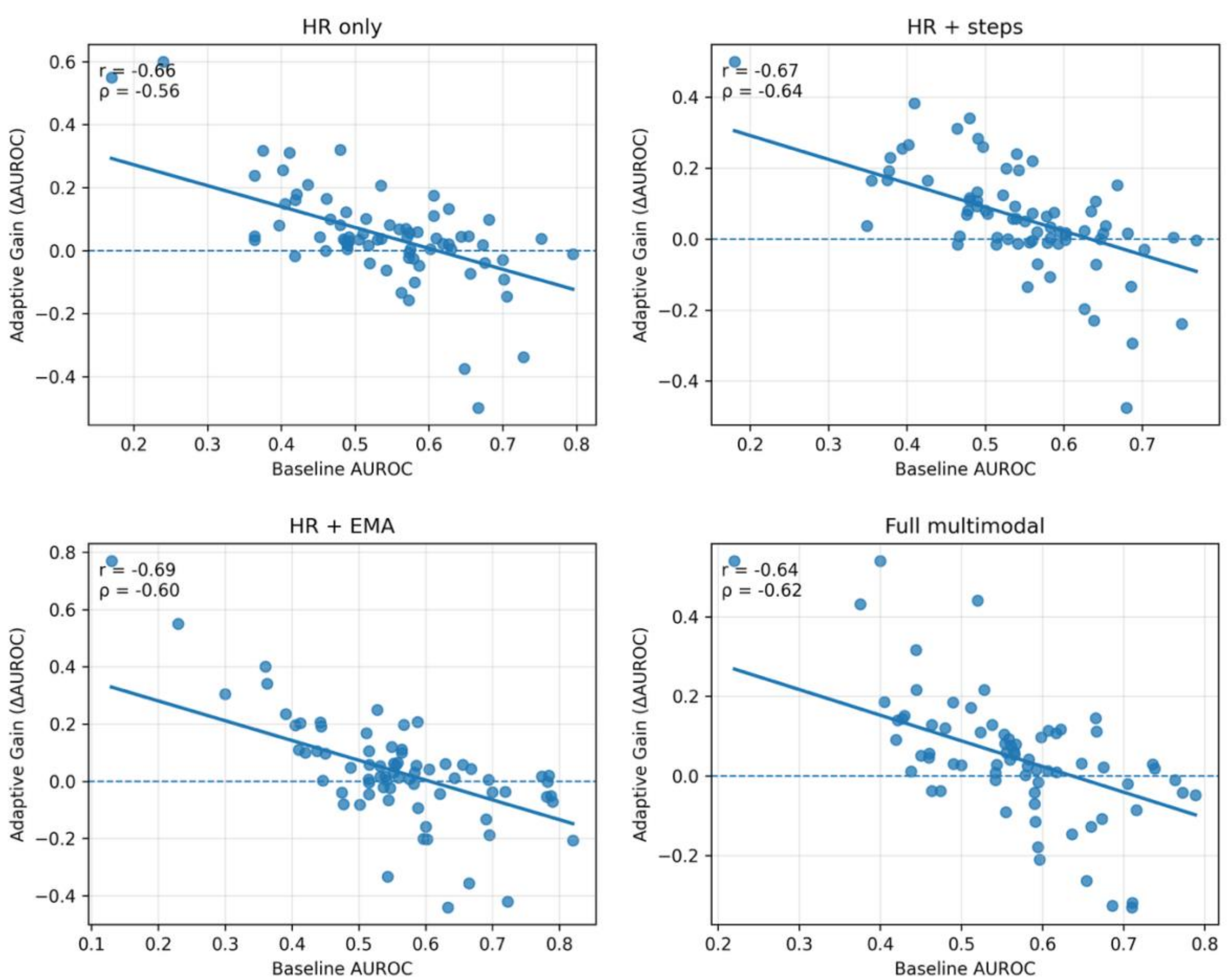

**Figure 2. Relationship between baseline performance and adaptive gain across modalities.**
Scatter plots show the relationship between baseline (random sampling) AUROC and participant-level adaptive gain for each modality configuration: (top-left) HR only, (top-right) HR + steps, (bottom-left) HR + EMA, and (bottom-right) full multimodal. Each point represents a participant. Solid lines indicate linear regression fits, and dashed horizontal lines denote zero gain. r and ρ correlation coefficients are reported within each panel. Across all modalities, a consistent negative association is observed.

For AUROC, this relationship was strong ($r = -0.67$, $p < 1e-30$; $\rho = -0.62$), indicating that participants with weaker baseline models benefited most from adaptive sensing. Participants who already had higher baseline performance tended to benefit less, and in some cases their performance worsened. A similar but weaker trend was observed for F1 score ($r = -0.17$, $p = 0.0015$; $\rho = -0.25$).

## Participant-level gains reveal substantial heterogeneity

To better understand individual-level effects, we analyzed the distribution of adaptive gains across participants (Figure 3).

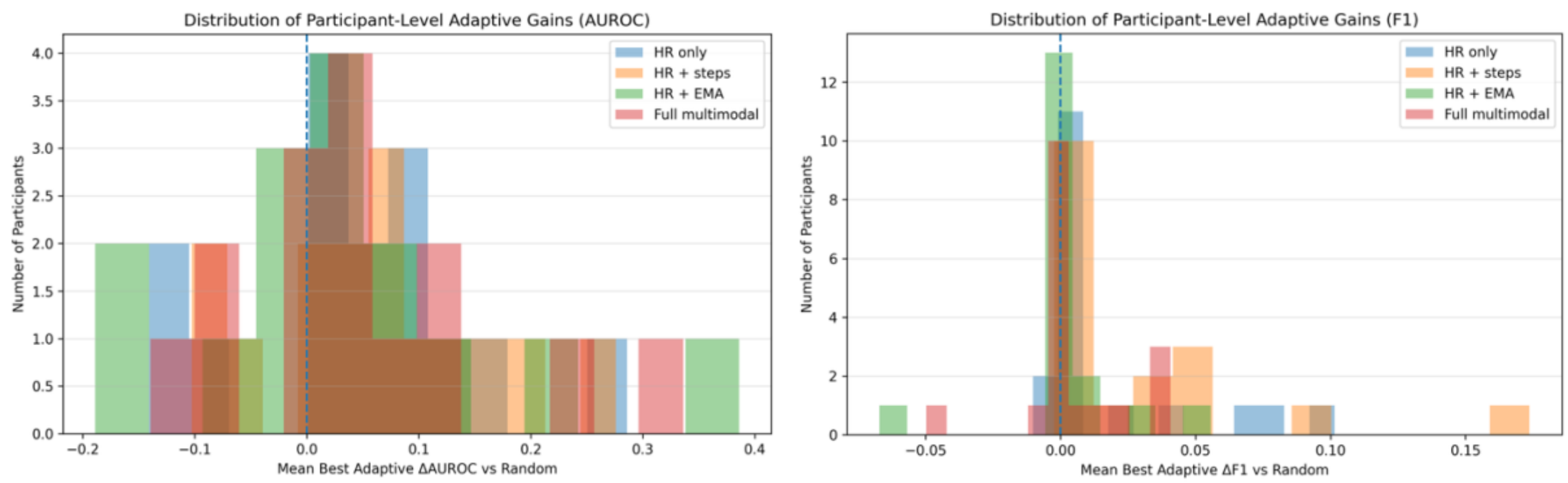

**Figure 3. Distribution of participant-level adaptive gains.**
(Left) Distribution of adaptive gains (AUROC) across participants for all modality configurations. Most participants show positive gains, with a right-skewed distribution indicating improvements for a subset of individuals. (Right) Distribution of adaptive gains in F1 score, showing smaller and more variable improvements centered around zero, reflecting reduced consistency across participants. This discrepancy likely arises because AUROC reflects ranking performance, whereas F1 depends on a fixed threshold that makes it more sensitive to class imbalance and calibration effects under subsampling.

For AUROC, most modality configurations showed a right-skewed distribution, with a subset of participants achieving substantial improvements (up to 0.30–0.35), while others experienced minimal or negative gains. Full multimodal models improved performance in 10/15 participants, while HR-only and HR+steps configurations showed improvements in 12/15 and 11/15 participants, respectively. Gains in F1 score were more modest and centered around zero, with fewer participants benefiting consistently (e.g., 7/18 improved under full multimodal settings). This suggests that adaptive sensing more reliably improves ranking performance (AUROC) than threshold-dependent metrics such as F1 score. This indicates significant inter-individual variability that reinforces the need for personalized sensing strategies.

## Adaptive sampling strategies outperform random selection under measurement constraints

We next evaluated model performance under varying measurement budgets using different sampling strategies (Figure 4).

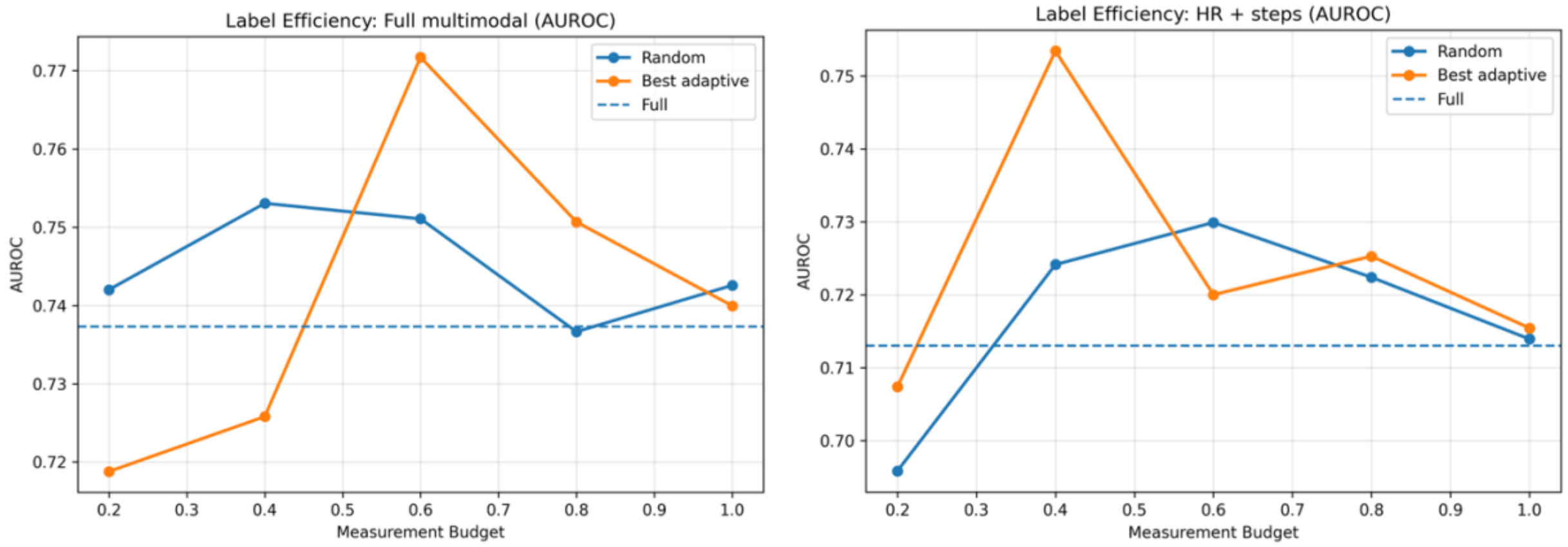

**Figure 4. Label efficiency under adaptive sensing strategies.**
Model performance (AUROC) as a function of measurement budget under random and adaptive sampling strategies. (Left) Full multimodal configuration. (Right) HR + steps configuration. Adaptive strategies consistently outperform random sampling, particularly at lower measurement budgets. It shows improved data efficiency and more effective use of limited observations.

Across budgets (20%–100%), uncertainty-based and hybrid selection policies most consistently outperformed random selection, particularly under lower data budgets. The magnitude of improvement varied by participant and modality, but gains were most pronounced at lower budgets, where selective measurement is most critical. Temporal variants occasionally improved performance by preserving coverage across different periods of the monitoring timeline, although gains varied across participants and modalities. Adaptive data selection can maintain or even improve predictive performance while reducing measurement burden, which is important for real-world deployment

### Multimodal integration does not uniformly improve adaptive gain

Although multimodal models generally achieved higher absolute performance, they did not always yield the largest adaptive gains. In several cases, simpler configurations (e.g., HR-only or HR+steps) achieved comparable or greater improvements under adaptive sampling. This shows the benefit of additional modalities depends on how effectively they contribute to informative sample selection, rather than their standalone predictive power.

## Discussion

In this study, we evaluated adaptive data selection strategies for wearable health prediction across multiple sensing configurations, participants, and data budgets. Our results show that adaptive data selection is not uniformly beneficial but instead provides selective and context-dependent gains. In particular, improvements were strongly associated with baseline model performance, with the largest gains observed in low-performing settings. Across modalities, model-guided selection of training windows improved predictive performance under constrained data budgets relative to random window selection.

A central finding of this study is the strong inverse relationship between baseline performance and adaptive gain (Figure 1). Participants with weak baseline models exhibited substantial improvements under adaptive data selection, whereas those with strong baselines showed limited or even negative gains. This suggests that selectively retaining informative training windows is most useful when the underlying prediction problem is difficult or when discriminative signal is limited. In contrast, when baseline performance is already high, prioritizing only a subset of windows may reduce training diversity and lead to diminishing returns. These findings indicate that adaptive data selection is most valuable in low-information or difficult prediction settings rather than as a universal replacement for random window selection.

We also observe substantial heterogeneity across participants (Figure 3), with some individuals benefiting considerably from adaptive strategies while others show minimal or inconsistent improvements. This variability reflects differences in physiological dynamics, behavioral patterns, and data quality, all of which influence model performance. Importantly, the majority of participants show improvements in AUROC, indicating that adaptive data selection more consistently improves ranking performance, even when gains in threshold-dependent metrics such as F1 are less stable. This emphasizes the need for personalized sensing strategies that account for individual variability rather than applying uniform data collection policies.

Our analysis further demonstrates that adaptive data selection can improve predictive performance under constrained data budgets (Figure 4). Uncertainty-based and hybrid selection policies consistently outperformed random window selection at lower budgets, suggesting that comparable performance can be achieved using fewer retained training windows. This has important implications for real-world wearable health systems, where reducing data collection burden is important for long-term adherence and scalability. By prioritizing informative training windows, adaptive data selection may improve efficiency without substantially compromising predictive performance.

Interestingly, multimodal integration did not uniformly translate into larger adaptive gains. Although multimodal models generally achieved higher absolute predictive performance, simpler configurations such as HR-only or HR+steps often exhibited comparable or greater gains under adaptive data selection. This suggests that the effectiveness of adaptive policies depends not only on modality richness, but also on how effectively a modality configuration supports identification of informative training windows. In highly redundant or information-rich settings, the marginal value of adaptive data selection may therefore be reduced.

These findings have several implications for wearable health system design. First, adaptive data selection should be viewed as a selective optimization strategy rather than a default modeling approach, with deployment guided by baseline model performance or predictive uncertainty. Second, personalization appears critical, as the benefits of adaptive selection vary substantially across individuals. Finally, adaptive policies may provide a practical pathway for reducing training data requirements while maintaining predictive performance, supporting more scalable and user-friendly digital health systems.

This study has several limitations. First, our analysis is based on retrospective simulations of adaptive data selection rather than prospective deployment in real-world systems. Future work is needed to evaluate whether these gains remain stable under live adaptive collection settings. Second, although we evaluated multiple sensing configurations and selection policies, alternative modeling approaches and policy-learning frameworks, including reinforcement learning, may further improve selection efficiency. Finally, our experiments focused on short-term prediction tasks, and extending adaptive data selection to longer-term forecasting and intervention settings remains an important direction for future work.

Adaptive data selection provides meaningful improvements in wearable health prediction, but its effectiveness depends strongly on baseline model performance, modality configuration, and participant-specific characteristics. Across experiments, the largest gains were consistently observed in low-performing settings, while benefits diminished in stronger baseline regimes. These findings suggest that adaptive data selection should be applied selectively rather than uniformly and highlight the importance of personalized deployment strategies for efficient wearable health monitoring.

## Methods

### Study design and data sources

We conducted a longitudinal observational study using multimodal wearable and self-reported data collected from participants over approximately 30 days. Data sources included continuous wearable biosignals (heart rate and step count), EMA responses collected 4–6 times per day (stress, mood, and contextual information), and intermittent BP measurements obtained using a wrist-worn monitor (Omron HeartGuide), typically paired with EMA responses. All data streams were temporally aligned and aggregated into fixed-length windows for downstream modeling (Figure 5).

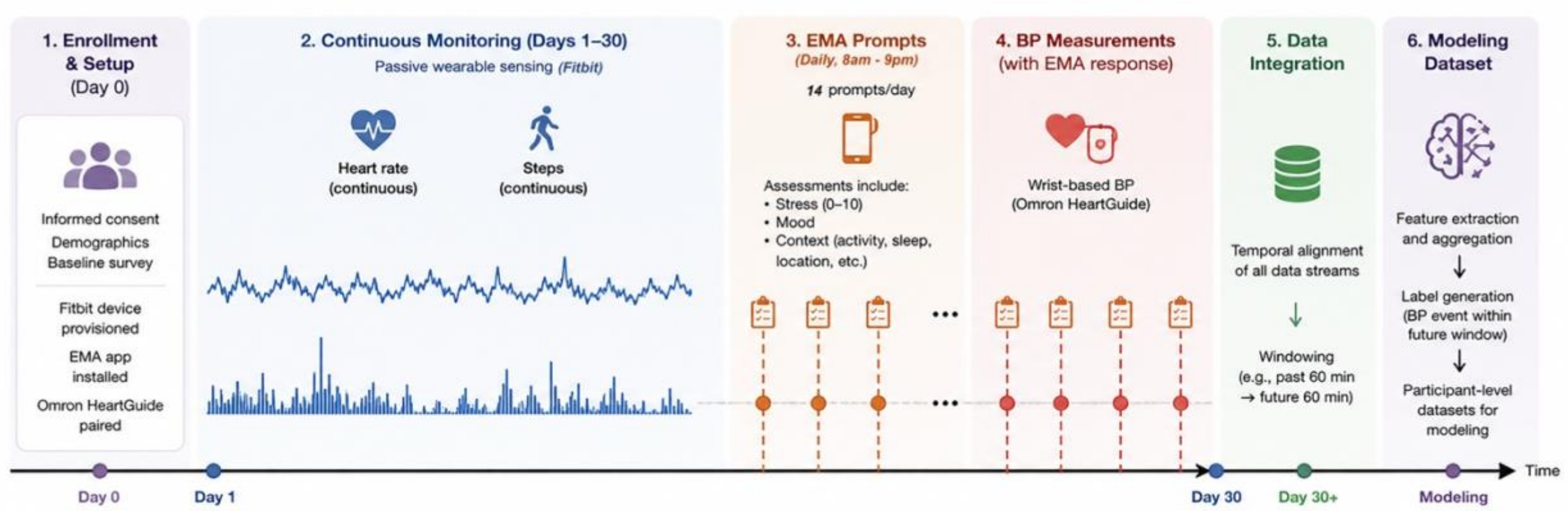

**Figure 5. Study timeline and multimodal data collection protocol.**
Participants were monitored longitudinally using wearable devices and EMA. Continuous physiological signals, including heart rate and physical activity, were passively collected over the monitoring period. EMA prompts were delivered multiple times per day to assess stress (0–10), mood, and contextual factors. Blood pressure measurements were obtained using a wrist-worn device in conjunction with EMA responses. All data streams were temporally aligned to construct analytical windows for downstream modeling.

**Data preprocessing, observation windows and measurement budget**

Raw data streams were temporally aligned and segmented into sliding windows, from which features were extracted across four modalities: physiological (heart rate statistics such as mean and variability), behavioral (step count and activity-derived measures), self-reported (EMA-based stress and contextual variables), and temporal context (e.g., time of day and day of week). Each window was assigned a binary label based on future blood pressure measurements, indicating whether an elevated BP event occurred within a predefined prediction horizon (e.g., 60 minutes), enabling a prospective prediction setting aligned with real-world monitoring. For each participant, all labeled windows constitute the candidate observation set available for model training. To simulate realistic sensing constraints, we introduced a measurement budget $b \in (0,1]$, representing the fraction of windows that can be selected for training. For example, a budget of 30% means that only 30% of all candidate windows are available for model training. Importantly, all sampling strategies operate under the same budget constraint, ensuring fair comparison.

**Training data budget simulation**

To simulate constrained data availability, we restricted the proportion of candidate training windows available for model training. Budgets ranged from 20% to 100% of all candidate windows in increments of 20%. For example, under a 20% budget, only 20% of available training windows were retained for model training. All selection policies operated under identical budget constraints to ensure fair comparison. Importantly, the budget constraint was applied only to the training set. Validation and test sets remained unchanged across all experiments.

**Adaptive data selection strategies**

We compared random window selection (baseline) with multiple adaptive data selection policies that differ in how candidate training windows are prioritized under a fixed data budget. Random window selection retains candidate training windows uniformly at random without using any model-derived information and serves as a baseline representing uninformed training window selection. In contrast, adaptive data selection strategies prioritize windows based on their expected informativeness for model learning. In this study, adaptive data selection refers to selecting a subset of temporal observation windows for model training under a fixed data budget. The sensing modalities, feature space, and model architecture remain unchanged across strategies; only the retained training windows differ. Uncertainty-based selection prioritizes windows where the model is least confident (e.g., predicted probability near 0.5), capturing ambiguous or difficult-to-classify cases. Risk-based selection prioritizes windows with the highest predicted probability of the target event (e.g., elevated BP), emphasizing clinically relevant or high-risk cases. Hybrid selection combines uncertainty and predicted risk scores to balance exploration of ambiguous cases with exploitation of high-risk cases. Temporal adaptive variants apply these same strategies within temporal strata (e.g., per day), ensuring temporal coverage and reducing over-concentration of selected windows within specific periods. For a given data budget $b$, each strategy retains the top-ranked proportion of candidate training windows according to its selection criterion. For example, under a 20% budget, only the top 20% of candidate training windows are retained for model training. All experiments use the same Random Forest architecture, feature space, and evaluation procedure. Differences in performance therefore arise solely from which training windows are retained under different selection policies.

### Model training and evaluation

We used Random Forest classifiers for all experiments to ensure consistency across data selection policies and modality configurations. Data were split temporally into training (70%), validation (15%), and test (15%) sets to prevent information leakage across time. Adaptive data selection policies were applied only to the training set, while validation and test sets remained unchanged across all experiments. Models were retrained using only the retained training windows selected under each policy and evaluated on the held-out test set. To quantify the benefit of adaptive data selection, we defined adaptive gain as the improvement in predictive performance relative to random window selection under the same data budget:

$$\Delta \text{Performance} = \text{Performance}_{\text{adaptive}} - \text{Performance}_{\text{random}}$$

Adaptive gain was evaluated for both AUROC and F1 score. For each participant, modality configuration, and budget, uncertainty-, risk-, hybrid-, and temporal-variant policies were evaluated independently. Participant-level adaptive gain analyses report the highest-performing adaptive policy relative to the corresponding random window selection baseline under the same budget. We report both participant-level (macro) and pooled (aggregate) performance across all participants. Importantly, all experiments used the same model architecture, feature space, and evaluation procedure. Differences in performance therefore arise solely from which training windows were retained under different data selection policies. The overall modeling and evaluation pipeline is illustrated in Figure 6.

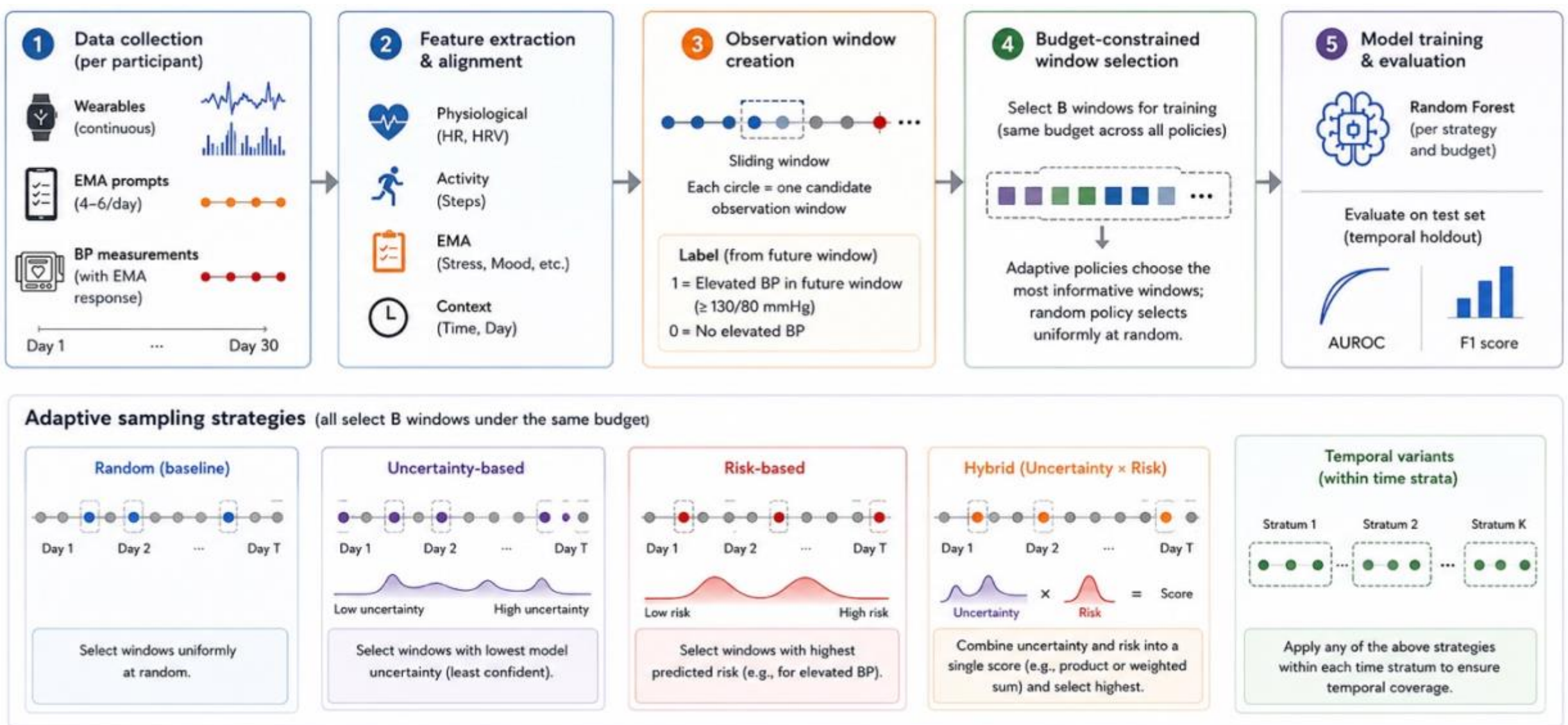

**Figure 6. Illustration of sampling strategies under a fixed measurement budget.**
Conceptual illustration of how different sampling strategies select observation windows from the same candidate set under a fixed budget. Random sampling selects windows uniformly. Uncertainty-based sampling prioritizes ambiguous samples where model confidence is low. Risk-based sampling selects windows with high predicted event probability. Hybrid sampling balances uncertainty and risk. Temporal variants apply these strategies within time strata (e.g., per day) to ensure temporal coverage. All strategies select the same number of windows, enabling fair comparison of their impact on model performance.

To characterize when adaptive sensing is beneficial, we analyzed adaptive gain as a function of baseline performance. For each participant and modality, baseline performance was defined as the performance of the model trained on randomly sampled data. We then computed the improvement achieved by the best adaptive strategy relative to this baseline. We quantified the relationship between baseline performance and adaptive gain using Pearson and Spearman correlation coefficients. Statistical significance was assessed using two-sided tests.

To assess heterogeneity across individuals, we computed participant-level adaptive gains by averaging improvements across budgets. We further quantified the proportion of participants exhibiting positive gains under each modality. Distributions of adaptive gains were visualized using histograms to characterize variability in outcomes. To evaluate data efficiency, we compared performance as a function of measurement budget across sampling strategies. For each modality, we identified the best-performing adaptive strategy at each budget and compared it to random sampling and full-data performance. These analyses quantify how effectively adaptive sensing recovers predictive performance under constrained data availability.

All analyses were conducted using Python (scikit-learn, pandas, and matplotlib). Correlation coefficients, summary statistics, and performance metrics were computed consistently across modalities and participants. No additional hyperparameter tuning or model selection was performed across experimental conditions to ensure fair comparison.

### Data Availability

A small, de-identified subset of the dataset is available in the project repository https://github.com/ucsfdigitalhealth/active_policy_learning_cardiomate to support reproducibility of the analysis pipelines on representative data. The full dataset (40 participants across two studies; approximately 4 weeks per participant) will be described and released in a companion dataset paper currently in preparation. All code used for data preprocessing and model development is available at the same repository.

## Ethics and Consent to Participate

This study was conducted in accordance with the Declaration of Helsinki and all relevant institutional guidelines and regulations. The research protocol was reviewed and approved by the University of Hawaiʻi Institutional Review Board (protocol #2023-00130) and received additional approval through the University of Hawaiʻi Data Governance Process (request #DGP 231116-4). All participants provided written informed consent on paper during the intake session before any study activities began. The consent process informed participants about data collection procedures, confidentiality protections, potential risks, and their right to withdraw at any time without penalty.

## Acknowledgments

AI-powered tools (e.g., ChatGPT) were used solely to assist with phrasing and grammatical refinement of an already complete draft. AI-based tools were also used in the creation of study design and workflow figures, in combination with manual editing using Adobe Photoshop.

## Competing Interest

The authors declare no competing financial or non-financial interests.